\documentclass[sigconf]{acmart}
\renewcommand\footnotetextcopyrightpermission[1]{} 

\usepackage{booktabs} 
\usepackage[linesnumbered,ruled]{algorithm2e}
\usepackage{subcaption}
\usepackage{hyperref}
\usepackage{cleveref}
\def\BibTeX{{\rm B\kern-.05em{\sc i\kern-.025em b}\kern-.08emT\kern-.1667em\lower.7ex\hbox{E}\kern-.125emX}}

 \copyrightyear{2019}
\acmYear{2019}
\setcopyright{acmlicensed}
\acmConference[]{}{}{}
\acmBooktitle{}
\acmPrice{}
\acmDOI{}
\acmISBN{}

\begin{document}

\title{Anomaly Detection for an E-commerce Pricing System}

\author{Jagdish Ramakrishnan}
\affiliation{%
  \institution{Walmart Labs}
  \streetaddress{850 Cherry Ave}
  \city{San Bruno}
  \state{CA}
  \postcode{94066}
}
\email{jramakrishnan@walmartlabs.com}
\author{Elham Shaabani}
\affiliation{%
  \institution{Walmart Labs}
  \streetaddress{850 Cherry Ave}
  \city{San Bruno}
  \state{CA}
  \postcode{94066}
}
\email{Elham.Shaabani@walmartlabs.com}

\author{Chao Li}
\affiliation{%
  \institution{Walmart Labs}
  \streetaddress{850 Cherry Ave}
  \city{San Bruno}
  \state{CA}
  \postcode{94066}
}
\email{CLi0@walmart.com}

\author{M\'aty\'as A. Sustik}
\affiliation{%
  \institution{Walmart Labs}
  \streetaddress{850 Cherry Ave}
  \city{San Bruno}
  \state{CA}
  \postcode{94066}
}
\email{MSustik@walmartlabs.com}

%
%
%
%
%
%
%



\vspace{-10pt}

\begin{abstract}
Online retailers execute a very large number of price updates when compared to brick-and-mortar stores. Even a few mis-priced items can have a significant business impact and result in a loss of customer trust. Early detection of anomalies in an automated real-time fashion is an important part of such a pricing system. In this paper, we describe unsupervised and supervised anomaly detection approaches we developed and deployed for a large-scale online pricing system at Walmart. Our system detects anomalies both in batch and real-time streaming settings, and the items flagged are reviewed and actioned based on priority and business impact. We found that having the right architecture design was critical to facilitate model performance at scale, and business impact and speed were important factors influencing model selection, parameter choice, and prioritization in a production environment for a large-scale system. We conducted analyses on the performance of various approaches on a test set using real-world retail data and fully deployed our approach into production. We found that our approach was able to detect the most important anomalies with high precision. 
\end{abstract}

%
%


%

\maketitle

\section{Introduction}

Pricing plays a critical role in every consumer's purchase decision. With the rapid evolution of e-commerce and a growing need to offer consumers a seamless omni-channel (e.g., store and online) experience, it is becoming increasingly important to calculate and update prices of merchandise online dynamically to stay ahead of the competition. At Walmart, the online pricing algorithm is responsible for calculating the most suitable price for tens of millions of products on Walmart.com. The algorithm takes both external data, such as competitor prices, and internal data, such as distributor costs, marketplace prices, and Walmart store prices, as inputs to calculate the final price that meets business needs (e.g., top line and bottom line objectives). The calculation is carried out in real-time with large amounts of data, which includes more than tens of millions of item cost data points and marketplace data points per day at Walmart. Many of the data sources are prone to data errors and some of them are out of the company's control. Data errors could lead to incorrect price calculations that can result in profit and revenue losses. For example, an incorrectly entered cost update for an item could drive a corresponding change for the price of the item. Note that an incorrect price of a digitally distributed product such as a code for a computer game could trigger significant financial losses within minutes without recourse. In addition, incorrect prices could hurt Walmart's EDLP (Every Day Low Price) value proposition, expose the company to a legal risk related to price agreements with manufacturers, and erode customer trust.

One approach to anomaly detection is to detect an anomaly if an item's price is more than a few standard deviations from its average historical price. However, given that large fluctuations in price are common in an online setting, this approach results in many false positives. Furthermore, not only do we want to detect price anomalies, we also want to identify and correct data errors that are the root cause of a price anomaly. This includes item attributes such as cost and reference prices from other channels.

To address this challenge, we developed a machine learning-based anomaly detection system that uses both supervised and unsupervised models. We used many features including prices from other channels, binary features, categorical features, and their transformations. We deployed the system into production in both a batch and streaming setting and implemented a review process to generate labeled data in collaboration with an internal operations team. Although the system was built for detecting pricing anomalies, we believe that the insights learned from model training, testing, and tuning, system and architecture design, and prioritization framework are relevant in any application that requires real-time identification and correction of data quality issues for large amounts of data.

Compared to previously published work on anomaly detection, the novel contributions of this paper are: 
\begin{itemize}
\item \textbf{An anomaly detection approach for a large-scale online pricing system} - While there are numerous applications of anomaly detection~\cite{Singh2012}, including intrusion detection, fraud detection, and sensor networks, there are relatively few references on anomaly detection in a retail setting. To the best of our knowledge, this is the first paper documenting methodologies, model performance, and system architecture on anomaly detection for a large scale e-commerce pricing system.
\item \textbf{Features and transformations for improving model performance} - The choice of features played an important role in model performance. We used a variety of features as inputs to our models, including price-based, binary, categorical, hierarchical, and feature transformations. We found that log-based feature transformations were especially useful for our Gaussian-based and autoencoder models. 
\item \textbf{An approach to explain anomalies} - While our proposed Gaussian Naive Bayes baseline model does not perform as well as the other sophisticated models, it served as the basis for our approach to explain anomalies. We used the scores from the model together with rule-based logic to provide reasonable explanations for why items were detected as anomalies. The use of explanations played an important role in directing human review of the items. 
\item \textbf{A prioritization scheme based on business impact} - Due to the large number of potential anomalies, we needed to prioritize and focus our attention on the ones that had the highest business impact. This prioritization allowed us to select the most important anomalies for human review given a fixed budget of items that could be reviewed in a given period of time.  
\item \textbf{Supervised and unsupervised models for anomaly detection} - Unsupervised approaches formed the basis of our initial models used in production because we had very few and insufficient labeled anomaly instances. As more items were reviewed, we increased our anomaly-labeled instances and moved to supervised approaches that had superior performance. Moving forward, we believe a combination of both supervised and unsupervised models will work well by using existing labeled anomalies efficiently, and at the same time, using models of normal instances to detect anomalies even when they are very different from those in our labeled set. 
\item \textbf{An architecture for model performance at scale} - The best performing model may not be the most effective model. We had to balance model performance and speed in a production environment to select the most effective model for our use case of real-time anomaly detection. For example, in our streaming setting, we used the baseline model rather than much better performing models because speed was a critical factor. In the batch setting, however, model performance played an important role, and we used the highest performing models. 
\end{itemize}

\section{Related Work}

There is a significant amount of literature on anomaly detection methods, including density-based~\cite{Kim2011, Fu2011}, quantile-based~\cite{Scholkopf1999, Tax2004}, neighborhood-based~\cite{Breunig2000, Kriegel2008}, and projection-based methods~\cite{Liu2008, Pevn2016}. A majority of papers focus on unsupervised methods, although there is representative work on both semi-supervised and supervised anomaly detection approaches~\cite{Gornitz2013}. Tree-based methods, such as Isolation Forest~\cite{Liu2008}, are especially attractive because they scale well with large datasets and have fast prediction times. Furthermore, they work well with diverse features, such as discrete and continuous, and the features do not need to be normalized. For other methods such as one-class SVMs~\cite{Scholkopf1999} and k-NN~\cite{Angiulli2002}, however, training time and/or memory requirements can rapidly increase with the size of the training data.

There has been recent interest in neural network based models for anomaly detection, including autoencoders~\cite{Zhai2016} and generative models such as GANs~\cite{Zenati2018}. In ~\cite{Zhai2016}, the authors propose deep structured energy-based models, where the energy function is defined as the output of a neural network. One example of such a model is an autoencoder where the anomaly score is the sum of the squared errors between the input and the output. We evaluate such an autoencoder model in this paper and compare its performance with other models.  

Much of the literature also focuses on time-series anomaly detection approaches~\cite{Hyndman2015, Shipmon2017, Zhu2017, Ahmad2016}. There are three main classes of anomalies: contextual anomalies, anomalous subsequences, and anomalous time-series~\cite{Kandanaarachchi2018}. Contextual anomalies are single observations at a particular point in time that are significantly different from their expected value. Anomalous subsequences are subsequences within a time series that differ from other parts of the series. And, anomalous time-series refers to time-series that differ from a collection of time series. In this paper, we are mainly focused on contextual anomalies because we are interested in whether an item is an anomaly at a particular point in time.

The use of anomaly detection methods in production systems is prevalent among large technology companies, including Yahoo~\cite{Laptev2015}, Google~\cite{Shipmon2017}, Facebook~\cite{Laptev2018}, LinkedIn~\cite{LinkedIn2015}, Uber~\cite{Zhu2017}, and Twitter~\cite{Vallis2014, Twitter2015}. While there is some work on anomaly detection in retail~\cite{Sabhnani2005}, to the best of our knowledge, there does not seem to be references on anomaly detection for a large-scale pricing system as we have considered in this paper. There are blog posts by companies such as Anodot~\cite{Anodot}, but they do not describe models in detail specific to the pricing context. 

\section{Methodology}

In this section, we introduce the proposed anomaly detection system for Walmart's online pricing system. We first describe the features and their transformations that we used for our models. Next, we introduce the unsupervised and supervised models that we used. Finally, we describe the overall process and system architecture of the deployed models.

\subsection{Features}

We describe the various types of features that we extracted for our problem. Let $\mathbf{x}$ represent the feature vector, where the  $i$th feature is represented by $x_i$. Depending on the model we use, the set of features may change, but we will make it clear exactly which set of features we are using. 

\textbf{Price-based features}. In our data, we identify price-based features as those that have some notion of a price point. The most obvious ones are the current price, competitor prices, and prices from other channels such as store prices, but cost is also an example of a price-based feature as it is relatable to the price of an item. We denote the set of indices of the \emph{raw} price-based features by $\mathcal{P}$, i.e., $x_i$ is a raw price-based feature if $i \in \mathcal{P}$. We also use time series based features, whose indices are represented by the set $\mathcal{T}$, such as the average of the historical prices, its standard deviation, and the percentage price swing from yesterday's price. 

\textbf{Binary, categorical, and other numerical features}. We have binary features such as whether an item is part of a marketing promotion that impacts pricing or is part of a bundle. We represent the set of indices of binary features as $\mathcal{B}$. For categorical features, we have data such as the type of promotion and the type of pricing algorithm we use. We convert the categorical features into binary form for modeling purposes. The set of indices for categorical features is represented by $\mathcal{C}$. Other numerical features, whose indices we represent by the set $\mathcal{O}$, include inventory, average customer rating of the item, and number of units in the item for say a multi-pack item.

\textbf{Hierarchical features}. Additionally, we also have hierarchical based features, i.e., sub-category, category, department, super-department, and division to which an item belongs. Higher levels of hierarchy contain subsets of the lower levels, e.g., multiple sub-categories are part of a category. Items within a particular hierarchy level may exhibit different characteristics than others. For example, the electronics category may be selling products at a lower margin than the jewelry category. Each hierarchy feature is categorical and represented by an integer. We denote the set of indices of hierarchical features as $\mathcal{H}$. A one-hot encoding of all the hierarchical features can result in a very large number of features; this can be challenging to incorporate in models. For some models, we choose to use a one hot encoding of only a single hierarchy feature, e.g., the department level. We describe this in more detail when we introduce models in Sections \ref{sec:baseline} and \ref{sec:models}.

\textbf{Transformations of price-based features}. We use a variety of feature transformations as inputs for our models. Since the price and cost features come up often in our transformation, we use the notation $x_p$ and $x_c$ respectively to refer to them. One set of transformations we use are features that indicate how far the price is from the cost of the item. These include differences
\begin{equation*}
x_i - x_c, \qquad i \in \mathcal{P}, \quad x_i \neq x_c
\end{equation*}
and margins
\begin{equation*}
\frac{x_i - x_c}{x_c}, \qquad i \in \mathcal{P}, \quad x_i \neq x_c
\end{equation*}
with respect to the raw price-based features. Similarly, the same set of the above transformations can be applied using $x_p$ in place of $x_c$, where $i$ is over all elements in $\mathcal{P}$ except $x_c$ and $x_p$. We refer to all of the above transformations with respect to cost and price as price transformed features and represent the full set of indices for these features as $\mathcal{P}_T$. 

\begin{figure}[h]
  \centering
  \includegraphics[width=\linewidth]{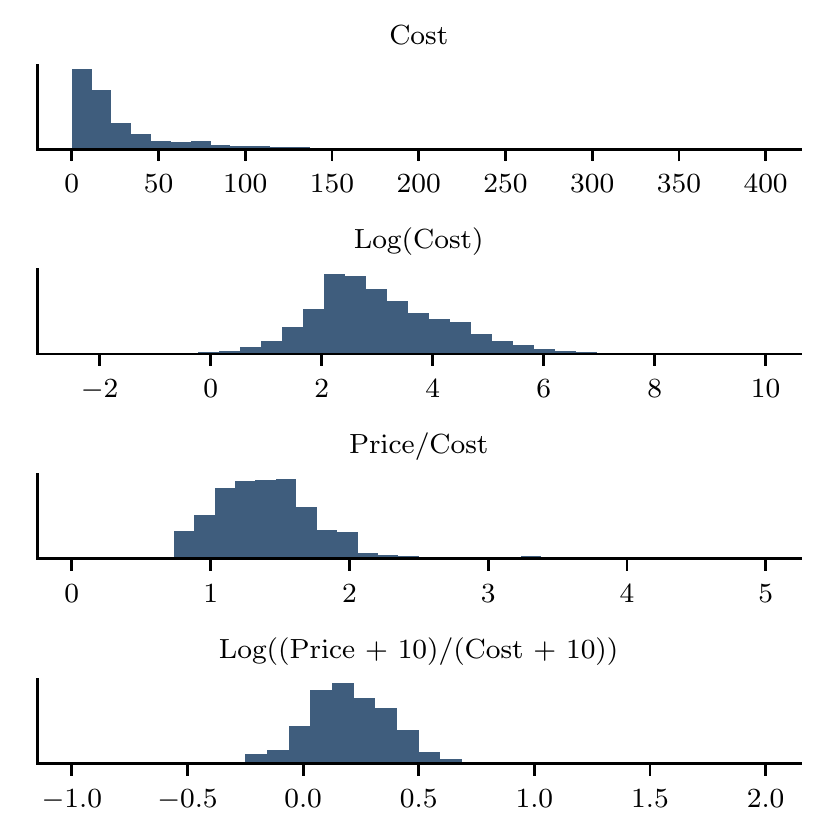}
  \caption{Log transformations of price-based features.}
  \Description{Log transformations of price-based features.}
  \label{feature_transform}
\end{figure}

Next, we discuss log transformed price features. For some models such as the autoencoder, we found that log based transformations are helpful in improving performance and speeding up training. This type of behavior for neural network models has been reported previously~\cite{Halder2018}. For ease in explanation, we denote the set $\mathcal{A}$ to be the indices of the six features that are the most important features in our pricing algorithm, e.g., price, cost, average historical price. Our baseline model makes use of these features. Using these features, we found log based transformations of the following form helped make the feature look more Gaussian:
\begin{equation}
\label{NB_transform}
\log \left( \frac{x_i + c_1}{x_c + c_1} \right) + c_2, \qquad i \in \mathcal{A}, \quad x_i \neq x_c,
\end{equation}
where $c_1$ and $c_2$ are appropriately chosen constants. Figure \ref{feature_transform} shows two examples of the above transformation. For these specific log transformations for features in $\mathcal{A}$ with $x_c$ in the denominator, we denote the log transformed feature indices by the set $\mathcal{A}_L$. We apply the same set of transformations using $x_p$ in the denominator, where $i$ is over all elements in $\mathcal{A}$ except when $x_i$ equals $x_c$ and $x_p$. Finally, the set $\mathcal{P}_L$ contains the indices of the full set of log based transformations including the indices in $\mathcal{P}_T$ and also the set $\mathcal{A}_L$. A summary of the different feature types, their notation, and feature count are provided in Table \ref{table:features}.

\begin{table}[ht]
\caption{Features used for models. AvgHistPrice is short for average historical price, IMU is the initial markup percentage or the margin percentage, i.e., (Price - Cost) / Cost, IsPromo is a binary feature indicating whether the item is part of a marketing promotion, and PromotionType is the type of such a promotion. } 
\centering 
\begin{tabular}{l l r l l} 
\hline\hline 
Feature Set & Notation & \# Features & Example \\ [0.5ex] 
\hline 
Raw Price & $\mathcal{P}$ & 17 & Price, Cost \\
Baseline Price & $\mathcal{A}$ & 6 & Price, Cost \\
Baseline Log Price & $\mathcal{A}_L$ & 5 & log(Price / Cost) \\
Time Series & $\mathcal{T}$ & 2 & AvgHistPrice \\
Transformed Price & $\mathcal{P}_T$ & 32 & IMU\\ 
Log Transformed & $\mathcal{P}_L$ & 39 & log(Price / Cost) \\
Hierarchy & $\mathcal{H}$ & 5 & SubCategoryId \\
Binary & $\mathcal{B}$ & 9 & IsPromo \\
Categorical & $\mathcal{C}$ & 3 & PromotionType \\
Other Numerical & $\mathcal{O}$ & 3 & Inventory \\ [1ex] 
\hline 
\end{tabular}
\label{table:features} 
\end{table}

\subsection{Gaussian Naive Bayes Baseline Model} 
\label{sec:baseline}

We describe the Gaussian Naive Bayes (GaussianNB) approach, which was a good starting point and a baseline for our anomaly detection models. We only use the log transformed versions of the main features used in our pricing algorithm, whose indices are described by the set $\mathcal{A}_L$. The basic idea of density-based anomaly detection models are that we build a probability distribution $p(\mathbf{x})$ of the normal class and if the density is below some threshold, we classify as an anomaly. The assumptions for GaussianNB are that the features are conditionally independent of the normal class, and the likelihood of the features are Gaussian. This would mean we have 
\begin{equation}
\label{NB_prod}
p(\mathbf{x}) = \prod_{i \in \mathcal{A}_L} p(x_i),
\end{equation} 
where the $p(x_i)$ is the likelihood corresponding to the feature $x_i$, and 
\begin{equation}
\label{NB_density}
p(x_i) = \frac{1}{\sqrt{2 \pi \sigma_i^2}} \exp{\frac{-(x_i - \mu_i)^2}{2 \sigma_i^2}},
\end{equation}
where the $\mu_i$ and $\sigma_i$ are the mean and the standard deviation of the $i$th feature, respectively. The choice of the threshold can then be selected from a validation set to get an ideal tradeoff between precision and recall.

Since items are part of a particular hierarchy level, e.g., a category, we can have a different model for each hierarchy level. Fitting models at a very low hierarchy level, e.g., at a subcategory level, have low bias but may overfit on the training set. On the other hand, fitting models at a higher hierarchy level may have a higher bias and underfit. We will explore the tradeoff between the various setups in the numerical experiments section.

\subsection{Explaining Anomalies}

While having the ability to predict anomalies is important, it is equally important to be able to guide a human reviewer to the cause of the anomalies. Given that there could be many possible reasons for an anomaly, we need to direct a reviewer to possible suspected issues. The advantage of the simple GaussianNB model is that we can use it to infer possible suspected issues. As before, we use only the feature indices represented by $\mathcal{A}_L$. We use transformations of form (\ref{NB_transform}), where the Cost feature $x_c$ is used in the denominator and all other features are used in the numerator. This results in a total of five features, each of which can give some indication of an issue with either the numerator feature or the denominator feature $x_c$. As mentioned earlier, using these log transformations makes features look more Gaussian, which the GaussianNB model would model well. These features provide information about how each price-based feature compares to the Cost of the item.

With GaussianNB, each feature can be assigned an anomaly score. To obtain the anomaly score, we take the log transformation of the density and multiply the resulting quantity by a constant. Now, from equations (\ref{NB_prod}) and (\ref{NB_density}), it can be seen that the anomaly score $A(\mathbf{x})$ is 
\begin{equation*}
A(\mathbf{x}) = \sum_{\{ i \in \mathcal{A}_L: A_i(x_i) \neq \text{NaN} \}} A_i(x_i) = \sum_{i \in \mathcal{A}_L} \frac{(x_i - \mu_i)^2}{\sigma_i^2},
\end{equation*}
where $A_i(x_i)$ is the anomaly score associated with the $i$th feature, and we define $A_i(x_i)$ to equal to NaN whenever the numerator feature in (\ref{NB_transform}) is missing. Now, we can choose a threshold $\epsilon$ so that if $A(\mathbf{x})$ is above $\epsilon$ we would predict an anomaly. We define $L[i]$ for $i \in \mathcal{A}_L$ to be the name of the numerator feature associated with the $i$th feature, e.g., for $\log((\text{Price} + c_1) / (\text{Cost} + c_1)) + c_2$ it would be "Price." Given the anomaly scores and their associated names, we would like to output a list of suspected issues, which we represent as $S(\mathbf{x})$. The detailed pseudocode and logic of the algorithm is provided in the supplemental Section \ref{explain_pseudo}.

\subsection{Beyond the Baseline Model.}
\label{sec:models}

We use four other approaches beyond the baseline GaussianNB model: Isolation Forest~\cite{Liu2008}, Autoencoder~\cite{Aggarwal2016}, Random Forest (RF)~\cite{Breiman2001}, and Gradient Boosting Machine (GBM)~\cite{friedman2001}. We also tried neighborhood-based approaches such as k-NN~\cite{Angiulli2002}, LOF~\cite{Breunig2000}, Fast ABOD~\cite{Kriegel2008} and quantile-based methods such as one-class SVMs~\cite{Schlkopf2001}; however, their training time or prediction time were too long for our scale. Isolation Forest and Autoencoder are unsupervised approaches, while RF and GBM are supervised approaches. We found tree-based approaches, i.e., Isolation Forest, RF, and GBM provided good performance and prediction times as we will see in the experiments section. For these approaches, we used the features from Table \ref{table:features} without any normalization. For the few categorical features, we used one-hot encoded features, and for the hierarchical features, we left them as label encoded.  The Autoencoder approach, on the other hand, required normalization and had much better performance when we used log transformed features. Further details are provided in the supplemental Section \ref{exp_details}. We used the sum of squared errors of the input vector and output vector as the anomaly score for the Autoencoder approach. For Isolation Forest, the anomaly score is the average path lengths of the tree. The supervised tree-based approaches used the probability of anomaly prediction as an anomaly score. Table \ref{table:models} summarizes the details of the various models and the number of features that they use. Note the Autoencoder approach uses fewer features than the tree-based models because it only uses log transformed features.

\begin{table}[ht]
\caption{Models.} 
\centering 
\begin{tabular}{l l r r} 
\hline\hline 
Approach & Type & \# Features \\ [0.5ex] 
\hline 
GaussianNB & Unsupervised & 5 \\
Isolation Forest & Unsupervised & 121 \\
Autoencoder & Unsupervised & 89 \\
GBM & Supervised & 121 \\ 
RF & Supervised & 121 \\ [1ex] 
\hline 
\end{tabular}
\label{table:models} 
\end{table}

\subsection{Threshold Selection}

For all the approaches, we vary the threshold $\epsilon$ and select the one that maximizes the standard $F_1$ score given by $2 \frac{\text{precision} \cdot \text{recall}}{\text{precision} + \text{recall}}$. We use cross-validation on the test set; we predict for each fold using the threshold selected from maximizing the $F_1$ score from the remaining folds. An alternative approach that we use in our production system is to choose the threshold that maximizes the recall at a minimum precision level, e.g., 0.80. 

\subsection{Prioritization Based on Business Impact}

With limited resources in investigating and fixing every anomaly, we needed to prioritize anomalies based on estimated business impact, or potential loss for Walmart, which we define as 
\begin{equation*}
\text{business\_impact} = \max\{\text{profit\_loss, foregone\_revenue}\}.
\end{equation*} 
Here, we assume profit\_loss to be loss caused by an incorrect low price while foregone\_revenue to be loss as a result of an incorrect high price. Given that there is no single data source to calculate business\_impact with 100\% accuracy, we decided to estimate these quantities by using
\begin{equation*}
\text{profit\_loss} = \max_i\{x_i - x_p\} \times \text{Inventory}, \qquad i \in \mathcal{A}, \quad x_i \neq x_p
\end{equation*}
and
\begin{equation*}
\text{foregone\_revenue} = \min_i\{x_i\} \times \text{Inventory}, \qquad i \in \mathcal{A},
\end{equation*}
where the $\max$ and the $\min$ above are taken over the features that are not missing, i.e., not equal to NaN. We then prioritize anomalies based on business\_impact.


\subsection{System Architecture}
\label{sec:arch}

\begin{figure}[h]
  \centering
  \includegraphics[width=\linewidth]{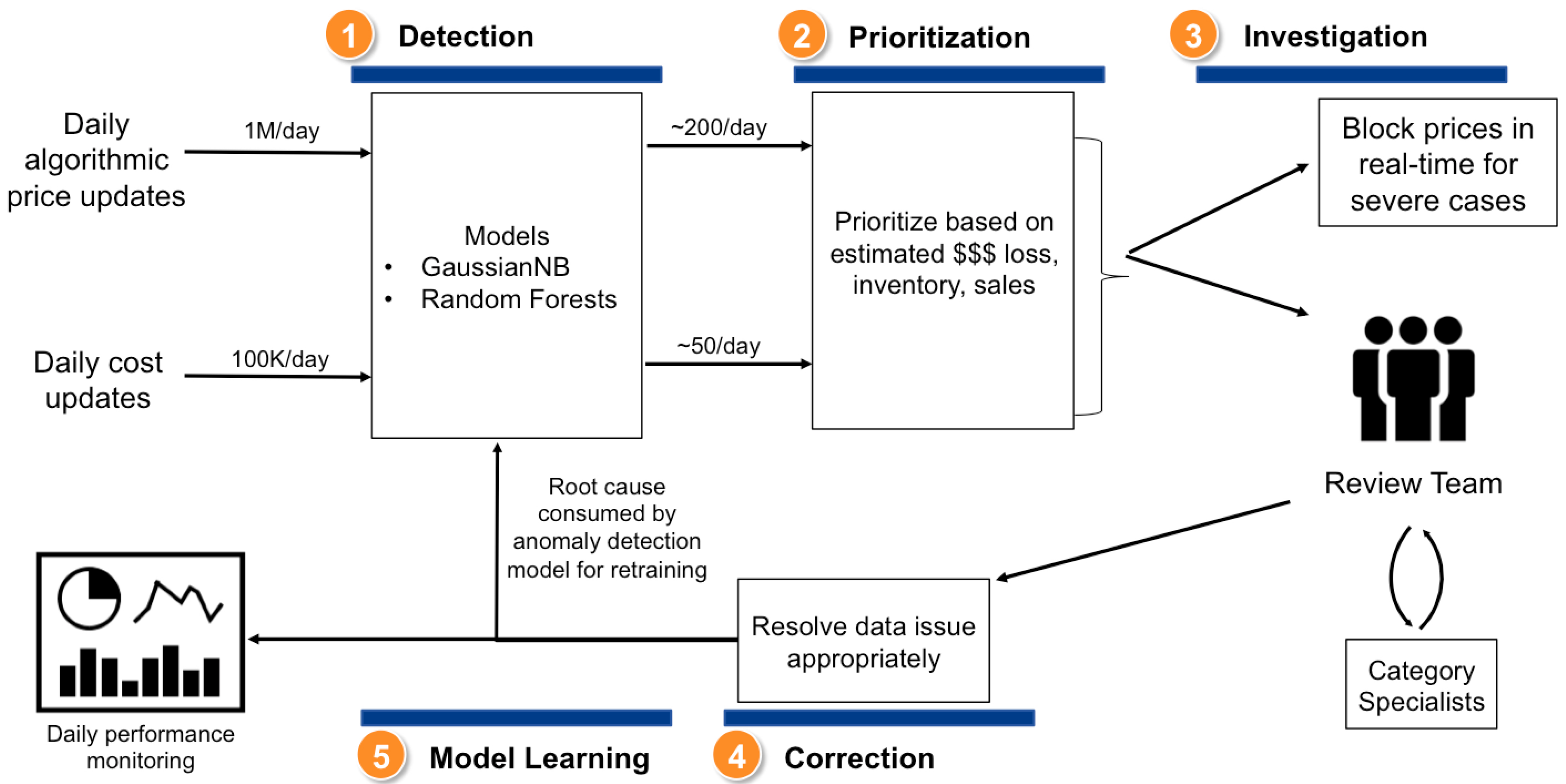}
  \caption{Overall system process.}
  \Description{Overall system process.}
  \label{figure:process}
\end{figure}

The overall system consists of detection, prioritization, investigation, correction, and model learning, as shown in Figure \ref{figure:process}. Based on more than 1M daily price updates and 100K cost updates, our anomaly detection models (e.g., GaussianNB, RF) predict anomalies and prioritize them based on business impact. The most severe anomalies that have high business impact are sent to a manual review team that has a capacity to review a fixed number anomalies daily. The reviewed anomalies are appropriately channeled to category specialists who correct the problem appropriately. Finally, the feedback obtained from these items are used as training data for our models.

We deployed our models through two different setups: a batch pipeline and a streaming pipeline. In the batch case, we had a daily job that applied our anomaly detection models on our entire product catalog with their current price. If items were anomalous and had a high business impact, they were sent for review. The alerted anomalies can be viewed through a web application by merchants and category specialists who can fix the data error. We set up a monitoring job that analyzed the progress of investigation and impact of each anomaly that was detected.

In the streaming setup, we block item prices in real-time for items with high business impact. Since we have millions of item prices that update in real-time, we take immediate action to block prices before they go live. Due to the scale of this system, it was crucial for us to ensure that predictions were made in less than a millisecond. This is one primary reason we are currently using the GaussianNB over other approaches for the streaming case. We are continuing to explore the use of more sophisticated models for our streaming pipeline within the speed constraints. The online pipeline uses Kafka, Flink, and Cassandra, and sends API requests to a real-time pricing API after which anomaly detection is applied prior to finalizing prices. If the prices are anomalous and have high priority, they are not updated, and an alert is generated for a category specialist to review.  

\begin{figure*}
 \centering
  \includegraphics[width=\textwidth]{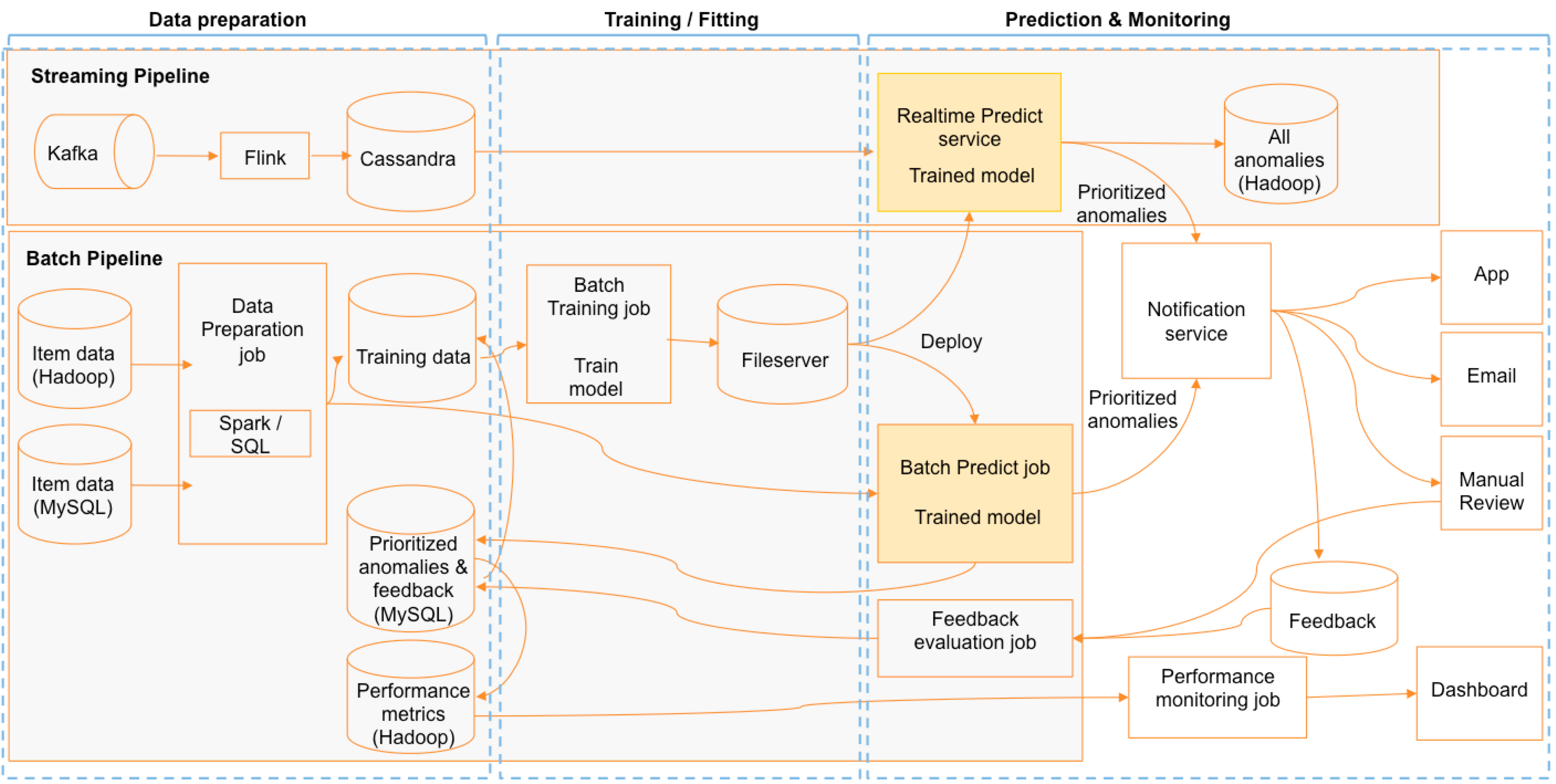}
  \caption{System architecture.}
  \Description{System architecture.}
  \label{figure:system}
\end{figure*}

Our anomaly detection models are trained in batch, and the model files are used appropriately by both the batch and streaming system. The models are stored as an object in a fileserver that is accessible by both systems. The training happens once a week. Data is collected from Hive / Hadoop and MySQL through Spark, and prepared appropriately for model training. For the batch system, since models are applied only once a day, the model file is loaded daily from the fileserver. The use of models in the streaming setup is more involved. In order for us to apply our models at scale, we have many compute nodes each of which are multi-threaded. We use the Flask~\cite{flask} micro web framework in Python. Since we do not want to reload the model file for every request due to the overhead and latency, we found the use of a TTL cache with expire time of a couple of hours to be a good solution. We used the Beaker package~\cite{beaker} for the TTL cache implementation. Another option to the TTL cache was to only load the model file when it is changed; this however is not a simple solution due to the load distribution among nodes and the multi-threaded architecture. The overall view of the system architecture is shown in Figure \ref{figure:system}. 

\section{Experiments}

We describe our experimental setup evaluating various anomaly detection approaches and their results. In the final part of this section, we describe how approaches were used in production and analyze the post-launch results. 

\subsection{Dataset and Data Preprocessing}

The dataset was created using real-world retail data and is highly class imbalanced. We collected anomaly data through two different ways. Before our business process was set up to have anomalies be manually reviewed through a support team, we set up a recording system where any time we noticed an anomaly, we manually recorded it in our system. Over the years, we have collected about two thirds of our anomaly data in this way. In the past year, we have set up a business process where a support team reviews items sent by our anomaly detection system. The ones marked as anomalies in this way are the remaining anomalies we have in our dataset. The total number of anomalies in our data is 2,137. 

For the normal data, we selected items from the Walmart catalog and assumed most must be correctly priced. We filtered out items that had extreme values for the main price-based features with indices in $\mathcal{A}$. We are aware that the normal instances in the dataset are contaminated with anomalies, but expect the contamination to be very small.

To create a training and test set for our performance evaluations, we randomly split the anomaly instances equally between train and test sets. For the normal instances, we randomly split the instances so that the percentage of anomalies in the test set is 0.1\%, which we think is reasonable. As we will see in the experimental results, the ratio of anomalies can effect the performance of the approaches. Table \ref{table:dataset} provides a summary of the train and test data. 

\begin{table}[ht]
\caption{Dataset.} 
\centering 
\begin{tabular}{l r r r} 
\hline\hline 
Class & Training Set & Test Set \\ [0.5ex] 
\hline 
Normal Instances & 4,627,747 & 1,066,932 \\
Anomaly Instances & 1,069 & 1,068 \\ [0.3ex]
\hline \\ [-2ex]
Total & 4,628,816 & 1,068,000 \\ [0.5ex] 
\hline 
\end{tabular}
\label{table:dataset} 
\end{table}

\subsection{Models at Different Hierarchy Levels}

We analyze the performance of GaussianNB models at different levels of hierarchy. For example, we could have a GaussianNB model for each subcategory, which would mean we have mean and standard deviation parameters for each subcategory. Fitting at lower hierarchy levels would mean more models and therefore could reduce bias but overfit and increase the variance. On the other hand, higher hierarchy levels would have fewer models and may result in a high bias. As we see in Table \ref{table:hierarchy}, fitting at the department level gives the right tradeoff and best performance in terms of the $F_1$ score. We used 5-fold cross-validation on the test set with stratified splits to select the threshold that results in the maximum $F_1$ score. The resulting predictions on all of the folds are used to calculate the reported $F_1$ score.

\begin{table}[ht]
\caption{Performance of GaussianNB models at different hierarchy levels. We use 5-fold cross validation with stratified splits. AUC refers to the area under the precision-recall curve.} 
\centering 
\begin{tabular}{l r r r r r} 
\hline\hline 
Hierarchy & \# Models & Precision & Recall & $F_1$ Score & AUC \\ [0.5ex] 
\hline 
SubCat& 2704 & \bf{0.3343} & 0.2210 & 0.2661 & \bf{0.1350} \\
Cat & 590 & 0.2827 & 0.2285 & 0.2527 & 0.1234 \\
Dep & 160 & 0.2894 & \bf{0.2303} & \bf{0.2565} & 0.1217 \\
SuperDep & 37 & 0.3051 & 0.2060 & 0.2459 & 0.1098 \\
Div & 7 & 0.2682 & 0.2247 & 0.2445 & 0.1000 \\ [1ex] 
\hline 
\end{tabular}
\label{table:hierarchy} 
\end{table}

\subsection{Performance Comparison of Models}

\begin{figure}[t!]
	\centering
  

  \subcaptionbox{Precision-Recall Curve\label{first-subfig}}{%
   \includegraphics[width=1\columnwidth]{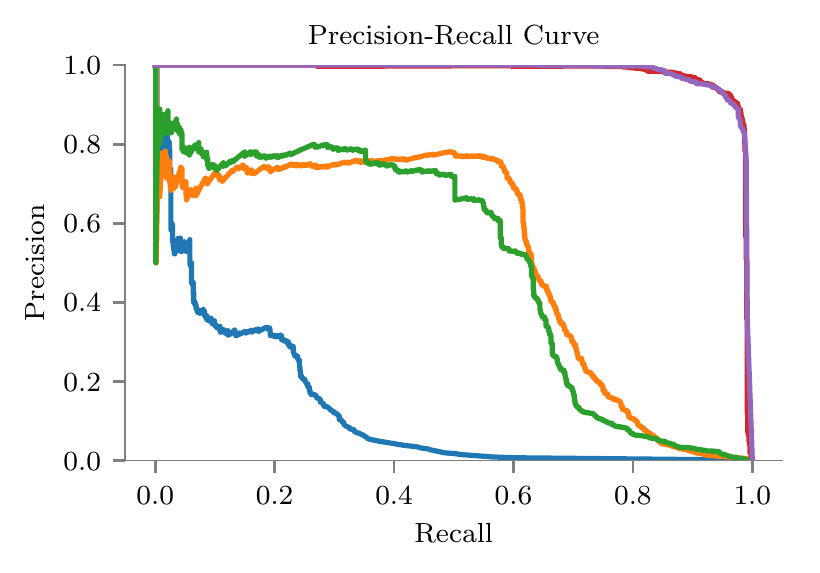} 
  }
  \subcaptionbox{Model Performance vs Percentage of Anomalies\label{second-subfig}}{%
   \includegraphics[width=1\columnwidth]{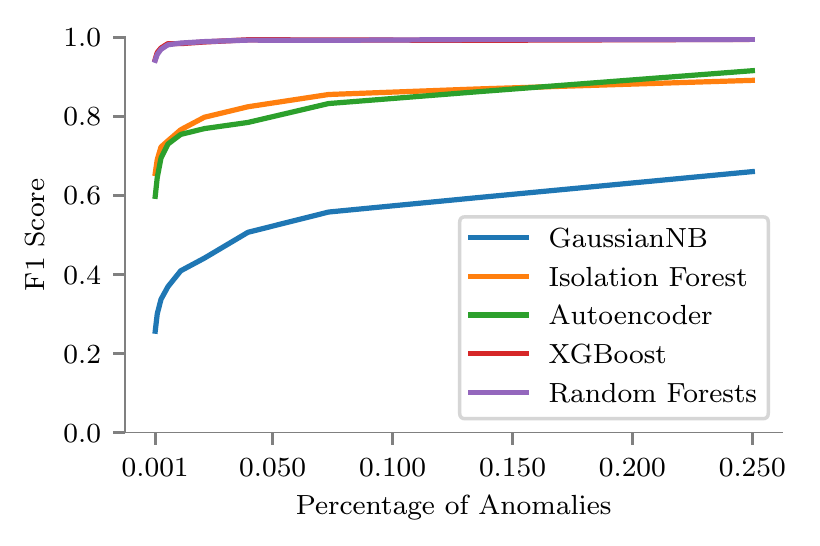}
  }

	\caption{Comparison of models. We use 5-fold cross validation with stratified splits. Plot (a) shows the precision-recall curves on the test data with 0.1\% anomalies. Plot (b) shows the $F_1$ score as a function of the percent anomalies in the test data; the percentage of anomalies on the plot ranges from 0.1\% to 25\%.}
  \Description{Comparison of models.}
  \label{figure:models}	
\end{figure}

We compare performance of the various models on the test set. We use 5-fold cross validation for the threshold selection, and report the precision, recall, $F_1$ score, and AUC from the results on the test set in Table \ref{table:model_results}. The supervised tree-based approaches, i.e., RF and GBM, both perform equally well and much better than other unsupervised approaches, e.g., Isolation Forest and Autoencoder. We also plot the precision-recall curves in Figure \ref{first-subfig}. These results assume that the test set has 0.1\% of anomalies as described in Table \ref{table:dataset}. There could be some amount of overfitting for the supervised approaches especially if the distribution of the test data does not reflect the actual distribution in reality. If there is not enough variety in the positive labeled data, the supervised approaches may not detect anomalies unseen in the labeled data while unsupervised approaches may be able to. 

\begin{table}[ht]
\caption{Performance of various anomaly detection models. AUC refers to the area under the precision-recall curve.} 
\centering 
\begin{tabular}{l r r r r r} 
\hline\hline 
Approach & Precision & Recall & $F_{1}$ Score & AUC \\ [0.5ex] 
\hline 
GaussianNB & 0.2894 & 0.2303 & 0.2565 & 0.1217 \\
Isolation Forest & 0.7555 & 0.5787 & 0.6554 & 0.5184 \\
Autoencoder & 0.6573 & 0.5478 & 0.5975 & 0.5008 \\
GBM & 0.9284 & \bf{0.9597} & \bf{0.9438} & 0.9810 \\ 
RF & \bf{0.9402} & 0.9429 & 0.9416 & \bf{0.9831} \\ [1ex] 
\hline 
\end{tabular}
\label{table:model_results} 
\end{table}

To see the effect of varying the percentage of anomalies, we undersampled the normal instances in the test set prior to the cross validation to obtain different anomaly percentages. We used 10 points for the percent anomalies from 0.1\% to 25\% in log space, and the results are shown in Figure \ref{second-subfig}. As we see in the plot, the lower the percentage of anomalies the lower the performance of all approaches. 

\subsection{Deployment in Production}

We describe deployment of models in production both for our batch and streaming system, and analyze post-launch results. For the streaming system, we use the GaussianNB approach. For the batch system, we used a combination of RF and GaussianNB approaches, which we describe in more detail below. The batch and streaming systems both use Algorithm \ref{explain_algo}, from the supplemental Section \ref{explain_pseudo}, for explaining anomalies with $\epsilon_s = \epsilon / 4$, where $\epsilon$ is the threshold used for selecting an anomaly. There were number of practical considerations that influenced our decisions for the deployed models. Below, we explain the details of the deployed system.

\textbf{Use progressive model deployment.} We adopted a progressive approach in model deployment to enable fast deployment and continual model improvement. Initially, we had a limited number of labeled anomaly data. Thus, we first deployed the baseline GaussianNB model because it was an unsupervised approach that did not require many anomaly instances. As we gathered more labeled data through manual review, we moved towards the supervised RF approach. We did this by integrating both the RF and GaussianNB models. The approach we took was to prioritize items that were identified as anomalies by both approaches, followed by those identified by RF but not GaussianNB, and finally, those identified by GaussianNB and not RF. After this, items were prioritized by business\_impact, potential loss, anomaly score from RF, and anomaly score from GaussianNB. The final prioritization was used to determine the top items that got sent for manual review and correction. The detailed steps and pseudocode are provided in Algorithm \ref{GNB_RF}.

\begin{algorithm}
    \SetKwInOut{Input}{Input}
    \SetKwInOut{Output}{Output}

    \underline{function combined\_predict} $(\mathbf{X}, L)$\;
    \Input{matrix $\mathbf{X}$ containing features for samples, list $L$ with issue names for getting suspected issues}
    \Output{vectors is\_anomaly, priority, suspected\_issues}
    is\_anomaly\_G, score\_G = GaussianNB.predict($\mathbf{X}$) \\
    is\_anomaly\_RF, score\_RF = RandomForest.predict($\mathbf{X}$) \\
    is\_anomaly = is\_anomaly\_G OR is\_anomaly\_RF \\
    suspected\_issues = get\_suspected\_issues($\mathbf{X}, L$) \\
    priority, business\_impact = business\_prioritization($\mathbf{X}$) \\
    Sort descending by is\_anomaly\_RF, is\_anomaly\_G, priority, business\_impact, score\_G, and score\_RF. \\
    \textbf{return} is\_anomaly, priority, suspected\_issues
    \caption{Combined GaussianNB and Random Forest predictions that was used in production.}
    \label{GNB_RF}
\end{algorithm}

\textbf{Choose fast prediction time for streaming system.} For our streaming system, we had strict time constraints to make a prediction. Due to the large number of daily price updates, the models had to make a prediction well within a millisecond. Due to this time constraint, we deployed the GaussianNB approach for our streaming anomaly detection API. Table \ref{table:time} shows the training and prediction times for the various approaches. The training time is not critical because we re-train the model only once a week. The batch prediction time is also not time critical since we only make predictions once a day for anomalies. However, the online prediction time of the different approaches informs us about the model possibilities for the streaming system, and we can see the GaussianNB approach is the fastest in the online setting. 

\begin{table}[ht]
\caption{Training and prediction times of anomaly detection models. We randomly sampled 1000 items from the test set with 25\% anomalies and reported the time from predicting them all at once (batch) and one-by-one (online). The prediction times are the average prediction time per item both for batch and online.} 
\centering 
\begin{tabular}{l r r r} 
\hline\hline 
Approach & Train & Batch & Online \\ 
 & time [s] & Prediction & Prediction \\  
 & & time [ms] & time [ms] \\ [0.5ex] 
\hline 
GaussianNB & 451.487 & 0.021 & 0.091 \\
Isolation Forest & 396.229 & 0.078 & 29.902 \\
Autoencoder & 6853.187 & 0.026 & 0.934 \\
GBM & 3138.834 & 0.005 & 0.169 \\ 
RF & 3794.588 & 0.321 & 215.925 \\ [1ex] 
\hline 
\end{tabular}
\label{table:time} 
\end{table}

\textbf{Weight precision over recall.} For both the batch and streaming system, we wanted to weight precision over recall to build trust in our anomaly detection system, especially initially. Because of this, we chose to maximize the $F_{0.1}$ score rather than the $F_1$ score for the GaussianNB approach, where the $F_\beta$ score is defined as $(1 + \beta^2) \frac{\text{precision} \cdot \text{recall}}{(\beta^2 \cdot \text{precision}) + \text{recall}}$. For the RF approach, we chose the threshold that maximizes the recall given a minimum precision of 80\%. We felt that using a system with at least 80\% precision would provide reasonable results. 

\textbf{Prioritize based on business impact.} For the streaming system, the main purpose of anomaly detection is to block prices that could have severe business impact. We chose to only block the prices that had the highest priority according to our prioritization logic. Once a price is blocked a category specialist is alerted through our web application, where they take actions to correct the issue or override with a manual price. For the batch system, we had a fixed capacity of alerts that could be reviewed by support team on a daily basis. 

\textbf{Results from production launch.} For the streaming system, while we have data on created alerts, the alerts were not thoroughly reviewed; the prices were automatically blocked due to the severity and only some alerts resulted in price corrections. In the batch system, however, every alert was reviewed through a support team and resolved as either a false positive or true positive and corrected appropriately. We analyzed the post-launch data for the batch system that we deployed. The approach used was the combined RF and GaussianNB approaches that we described earlier.

A total of 5,205 alerts were generated over two months, and only 1,625 alerts had a resolution. The typical total review time can varied between a week to a couple of weeks. Alerts were reviewed by a support team and appropriately directed to a team or a category specialist who is an expert on the specific item referenced. Once a category specialist determines whether there is an issue, they correct it if needed, and then, the alert is marked with a resolution.

As shown in Table \ref{table:production}, there were 836 false positive, resulting in a precision of 53.5\% among the alerts generated. It is not possible to measure recall because we do not actually know about existing anomalies. The actual precision of 53.5\% is significantly below the desired 80\% precision. In order to understand further if there was a bug or a systematic problem with our deployed approach, we conducted error analysis on on 100 randomly sampled alerts out of the 756 that were marked false positives. Our second review found that about 49\% of the items that were marked false positives were actually not false positives, and there was a systematic issue with the labeling. From category specialist point of view, if an item has a correct price and cost, everything is fine. However, for many items, even with a correct price and cost, there may be issues with other item data that could impact prices in the future. Indeed, our review team was marking items detected by our system with incorrect competitor prices as false positives. We believe these are not false positives, and our models were designed to catch these types of issues. If we adjust for this systematic error by generalizing the 49\% error rate over the fully reviewed set, this could change the precision from 53.5\% to 76.2\%, bringing it much closer to the 80\% desired precision. The original and adjusted numbers are reported in Table \ref{table:production}. Besides addressing this competitor price issue, we are working on ways to improve our manual review process, e.g., have multiple reviews for a subset of items for quality assurance. 

\begin{table}[ht]
\caption{Results from production launch. FP refers to the number of False Positives, i.e., number of predictions that were not actually anomalies. } 
\centering 
\begin{tabular}{l r r r r r} 
\hline\hline 
 & \# Alerts & \# Reviewed & \# FP & Precision \\ [0.5ex] 
\hline 
Original & 5,205 & 1,625 & 756 & 53.5\% \\
Adjusted & 5,205 & 1,625 & 386 & 76.2\% \\ [1ex] 
\hline 
\end{tabular}
\label{table:production} 
\end{table}

\section{Conclusion}

We proposed an anomaly detection framework for Walmart's online pricing system. Our models were able to detect the most important anomalies effectively by finding mis-priced items and incorrect data inputs to our pricing algorithm. Besides detecting anomalies, we developed an approach that relies on the anomaly scores from a density model to explain the anomalies detected. In order to concentrate on the most important anomalies to review and in turn further gather labeled data for our models, we used estimated business impact and other pertinent item information to prioritize the anomalies. We trained and evaluated various unsupervised and supervised approaches using real-world retail data. After selecting the appropriate models, we successfully deployed our approaches in production and achieved the desired precision of detection.

For future work, we can explore methods that systematically incorporate overlapping hierarchy levels such as done in~\cite{Dudik2007} through a maximum entropy formulation; in this paper, we considered hierarchical-based features by either using them as label encoded or as a one-hot vector. We can further consider more sophisticated learning-based models for explaining anomalies, such as~\cite{Siddiqui2019}. Another idea is to explore more extensive time series features such as the ones provided in~\cite{Fulcher2014}. Collection of reliable labeled data is an on-going challenge. Due to the complex nature of anomaly detection, human review is not foolproof. In addition to anomaly review by a dedicated team, we plan to explore other options such as crowd sourcing, to improve the accuracy of our data labeling process. 

\section{Acknowledgements}
We would like to thank the entire Walmart Smart Pricing team for their contributions to this project. We thank Marcus Csaky, Varun Bahl, Brian Seaman, and Zach Dennett for being very supportive of this project and for providing feedback on the paper. We thank Tracy Phung for her suggestions and work on training data and alerts generation. We thank Ravi Ganti for helpful discussions and suggestions about the autoencoder model. We thank Andrew Torson, Abhiraj Butala, and Vikrant Goel for suggesting the use of a TTL cache for the streaming pipeline. We thank Victor Oleinikov and Paulo Tarasiuk for work on alerting anomalies through a web application, Kevin Shah for his work on reviewer feedback, and Vaishnavi Ravisankar for her work on performance monitoring. 

\bibliographystyle{ACM-Reference-Format}
\bibliography{bibliography}

%
\newpage
\appendix

\section{Supplementary Information}

Implementation of approaches can be obtained at 

\noindent\href{https://github.com/walmartlabs/anomaly-detection-walmart}{https://github.com/walmartlabs/anomaly-detection-walmart}.

\subsection{Python Packages}

We used the pyod python package~\cite{zhao2019pyod} for the Autoencoder approach. For the Isolation forest, RF, and GBM implementations, we used the scikit-learn package~\cite{scikit-learn}. For GBM, we use the scikit-learn API for XGBoost~\cite{Chen2016}. For the GaussianNB approach, we wrote custom code using the numpy package~\cite{numpy}.

\subsection{Experiment Details}
\label{exp_details}

For the GaussianNB approach, we fit models on the department level, which meant we had a separate GaussianNB model for each department. There were 160 departments in our training and test data and therefore models. For some predictions, an item had a new department label that was unseen in the training data; for these cases, we used a GaussianNB model that was trained on the entire training data with data from all departments. We ensured that every model had at least 9 samples for fitting; if not, we fit the model at one level higher in the hierarchy, i.e., super-department. We ensured a minimum standard deviation of 0.01 by clipping it if it went below 0.01. 

For Isolation Forest, we used 100 estimators, 5\% of the training set as the number of samples, and 10\% of the features to train each base estimator. Initially, we used a fixed number of samples to train estimators, e.g., 512, but we found that the performance was severely impacted due to the low number of samples; 5\% of the training set (roughly 200K samples) worked well for us. For the Autoencoder model, we used 64, 32, 32, and 64 units for the encoder and decoder hidden layers respectively. We used ReLU activations for all hidden layers and a tanh activation for the output layer. We standardized the data prior to feeding to the input layer. We used mean squared error as the loss, i.e., the anomaly score, 512 for the batch size, 100 epochs to train the network, and the Adam optimizer~\cite{adam2014}. We used a 0.2 dropout rate, and 0.1 for the regularization strength of a activity\_regularizer on every layer. Most of these choices were defaults from the pyod package~\cite{zhao2019pyod}. We used 100 estimators and a max depth of 5 for GBM, and we used 400 estimators and a max depth of 80 for RF. 

\subsection{Computing Resources}

For all comparisons of the approaches, we requested the same cloud computing resources for fitting and prediction. We used a single node with 5 CPU cores and 45GB of RAM.

\subsection{Approach for Explaining Anomalies}
\label{explain_pseudo}

We describe the details of the logic used for explaining anomalies. The pseudocode is provided in Algorithm \ref{explain_algo}. The first loop (lines 4 - 10) collects all features in the suspected issues list $S(\mathbf{x})$ whose anomaly score have more than a given threshold $\epsilon_s$. These features are highly likely to be anomalies. Note this threshold $\epsilon_s$ is different from the threshold $\epsilon$ described earlier that determines whether or not we have an anomaly. Next, in the first if statement (lines 11 - 15), as long as we have enough features that do not have NaN, if the number of suspected issues with large anomaly scores is greater than 1, we believe the Cost is the issue; otherwise, we can infer Cost is not an issue. The intuition behind the approach is that since the Cost in the denominator of every feature, we can infer there is something wrong with it if the anomaly score from multiple features is large. If only a single feature has a large anomaly score, we can at best infer that either the feature in the numerator or the Cost is an anomaly; in this case, we provide both features as an explanation. In the final if statement (lines 16 - 22), we handle the case when Cost is a suspected issue. In this case, if Cost is a suspected issue and if there are any features that do not differ very much from Cost, i.e., $A_i(x_i) < \epsilon_s$, we would also mark that feature as an issue. As an example, consider 2 features, where Price is \$1 and a competitor price is \$100, that we apply the log transformations with a Cost of \$1 in the denominator. The competitor price feature will result in a high anomaly score, which means we will infer the competitor price and Cost as issues. Since Cost is an issue so is Price because it is not far from it. 

\begin{algorithm}
    \SetKwInOut{Input}{Input}
    \SetKwInOut{Output}{Output}

    \underline{function get\_suspected\_issues} $(\{x_i, A_i(x_i), L[i] \text{ for } i \in \mathcal{A}_L \})$\;
    \Input{features $x_i$, $A_i(x_i), L[i]$ for $i \in \mathcal{A}_L$, threshold $\epsilon_s$}
    \Output{list of suspected issues, $S(\mathbf{x})$}
    $S(\mathbf{x}) = []$ \\
    $\text{num\_not\_null} = 0$ \\
    \For{$i \in \mathcal{A}_L$}
    {
    \uIf{$A_i(x_i) \geq \epsilon_s$}
    {
    $S(\mathbf{x})$.append(L[i])
    }
    \ElseIf{$A_i(x_i) \neq \text{NaN}$}
    {
    num\_not\_null += 1
    }
    }
    \uIf{$\text{num\_not\_null} \leq 2$}
      {
     	 $S(\mathbf{x})$.append("Cost")
      }
    \ElseIf{$\text{len}(S(\mathbf{x})) > 1$}
    {
    	$S(\mathbf{x}) = [\text{"Cost"}]$
    }
    \If{$\text{"Cost" in } S(\mathbf{x})$}
    {
    \For{$i \in \mathcal{A}_L$}
    {
    \If{$A_i(x_i) \neq \text{NaN}$ and $A_i(x_i) < \epsilon_s$}
    {
    $S(\mathbf{x})$.append(L[i])
    }
    }
    }
    \textbf{return} $S(\mathbf{x})$
    \caption{Rule-based approach to predict suspected issues from GaussianNB anomaly scores}
    \label{explain_algo}
\end{algorithm}




\end{document}